\documentclass{article}

\usepackage{graphicx}
\usepackage[utf8x]{inputenc}
\usepackage{amsmath}
\usepackage{amsfonts}
\usepackage{color}
\usepackage{graphicx}
\usepackage{natbib}

\newtheorem{defn}{Definition}

\newtheorem{ex}{Example}

\begin{document}

\title{A Note on Data Simulations for Voting by Evaluation}



\author{Antoine Rolland\footnote{A. Rolland (corresponding author) 
 ERIC EA 3083, Universit\'e de Lyon, Universit\'e Lumi\`ere Lyon 2, 5 Pierre Mend\`es France, 69596 Bron Cedex, France,
              {antoine.rolland@univ-lyon2.fr}        }         \and
        Jean-Baptiste Aubin\footnote{Universit\'e de Lyon, INSA de Lyon, CNRS UMR 5208, Institut Camille Jordan,
F-69621 Villeurbanne, France, {jean-baptiste.aubin@insa-lyon.fr@insa-lyon.fr}} \and
        Ir\`ene Gannaz\footnote{Universit\'e de Lyon, INSA de Lyon, CNRS UMR 5208, Institut Camille Jordan,
F-69621 Villeurbanne, France, {irene.gannaz@insa-lyon.fr@insa-lyon.fr}} \and
        Samuela Leoni\footnote{Universit\'e de Lyon, INSA de Lyon, CNRS UMR 5208, Institut Camille Jordan,
F-69621 Villeurbanne, France, {samuela.leoni@insa-lyon.fr@insa-lyon.fr}} 
}


\date{31 march 2021}

\maketitle

\begin{abstract}
Voting rules based on evaluation inputs rather than preference orders have been recently proposed, like majority judgement, range voting or approval voting. Traditionally, probabilistic analysis of voting rules supposes the use of simulation models to generate preferences data, like the Impartial Culture (IC) or Impartial and Anonymous Culture (IAC) models. But these simulation models are not suitable for the analysis of evaluation-based voting rules as they generate preference orders instead of the needed evaluations. We propose in this paper several simulation models for generating evaluation-based voting inputs. These models, inspired by classical ones, are defined, tested and compared for recommendation purpose.
\end{abstract}

\section{Introduction }\label{sec:objectifs}

Voting rules can be seen as functions which aim at determining a winner in a set of candidates considering a set of voters. Both social and mathematical approaches consider positive or negative properties satisfied by a given voting rule as a matter of interest.
``Negative'' properties are usually called ``paradoxes'' even if they are only counter-intuitive. Studying the properties of voting rules can be done either in an axiomatic approach or in a probabilistic approach. The axiomatic approach supposes to determine which (set of) properties characterize a specific voting process, \emph{i.e.} which properties are to be observed, and which are not, via formal theorems. The probabilistic approach aims at determining whether these (intuitive or counter-intuitive) properties  are likely to be observed in real life, \emph{i.e.} determining the frequency of such properties using the voting rule. This pragmatic approach is based on simulations. Several models exist which have been well studied for years. One can refeer to \citet{Diss2020} for a recent state-of-the-art of simulating technics for a probabilistic approach of voting theory. \citet{Tideman2012,Tideman2014}, \citet{Plassman2014}, \citet{Green-Armytage2016} contain examples of simulation-based studies of voting rules. All these models are appropriate to study voting rules using preference orders on candidates as input. But other voting rules have been recently proposed based on evaluations and notations given by the voters about the candidates, therefore using preference intensities and not only preference orders. Classical simulation models are therefore inoperative for studying such voting process. 

We propose in this paper to investigate several methods to simulate evaluation/notation data for a simulation study of evaluation-based voting rules. 
We first introduce evaluation-based voting methods in section \ref{sec:evaluation.based.voting.methods}. We then introduce simulation methods analogous to Impartial Culture (IC) and Impartial and Anonymous Culture (IAC) frameworks in section \ref{sec:ICIAC} and methods based on spatial simulations in section \ref{sec:simuspatiale}. Comparison and evaluations of these methods are analysed in section \ref{sec:compmethods}, which ends with a discussion.

\section{Evaluation based methods}
\label{sec:evaluation.based.voting.methods}

Classical voting methods are based on preference rankings. These methods have a clear historical justification: before the computers, further preferences of voters couldn't be taken in consideration for practical reasons. Nevertheless, information loss during the process is an important issue. For example it's impossible to know if the ranking of a voter is a ``by default'' ranking (constituted of ``to reject'' candidates) or if there exists a subset of acceptable candidates. As a consequence, these classical methods based on rankings are vulnerable to numerous ``paradoxes'' and impossibility theorems. Hence, these methods can suffer from a lack of legitimacy. More specifically, the plurality with run-off voting procedure suffers from important paradoxes as lack of monotonicity, reinforcement, no-show, independence to irrelevant alternative or Condorcet winner paradox (see more details in \citet{Felsenthal2012}). Other methods based on evaluations exist. They are more nuanced and keep more information than the previous ones. Obtained results are promising and escape some limits of the classical methods.


Three of the most famous methods based on evaluations are the approval voting, the range voting and the majority judgement. 

The approval voting (see \citet{Brams2007} for a complete study) is maybe the most famous of these methods: each voter evaluates candidates on a scale of 2 gradings, which is the simplest possible scale.  The voter gives a 1 if the candidate is acceptable, else a 0. The voter can then votes for several candidates (even all of them), or none of them accordingly to his convictions. Note that this method is very simple to apply in practice.

The two others methods are based on more nuanced class of gradings, which can be continuous or on a discrete scale, and  aggregate the evaluations in a different way. With the range voting proposed by \citet{Smith00rangevoting}, the candidate with the highest average grade is the winner. With the majority judgement introduced by \citet{Balinski2007,Balinski2011}, the candidate with the highest median grade is the winner. Note that the tie-break situation is taken into account for example in \citet{Balinski2020} and \cite{Fabre2021}.

These three methods present a clear advantage with respect to the classical ones: they allow the possibility of a ``pacific revolution". If no candidate is ``satisfactory'' for the voters, one could cancel the election and start a new one with different candidates. For example, for the approval voting (respectively for the range voting or the majority judgement), a minimum percentage of approbation (respectively a minimum mean or median) should be reached to be elected . Moreover, these three methods are not vulnerable to ``paradoxes'' of non-monotonicity or independence to irrelevant alternatives.

It should be noticed that a common criticism against  these three methods is that they are sensitive to the so-called Condorcet-winner paradox and Condorcet-loser paradox. These two paradoxes are significant in the context of preference order based methods, as their meaning stands in the poor information available about the candidates (the rank of each candidate). They loose a large part of their legitimacy in the context of evaluation based methods, since the additional available information through a quantitative evaluation  modifies the picture of the situation. This new information can legitimate the election of a candidate who should be avoided at the sole sight of his one-on-one majority ballot confrontations. These two paradoxes can be seen as a measure of a voting system consistency with the one-round majority voting system. But  precisely evaluation-based voting rules should not be too ``consistent'' with the one-round majority voting system as one has access to information that is not taken into account by the majority voting system, as illustrated in the following example.
\begin{ex}
Suppose that 3 voters A,B and C have to choose one out of two candidates $x$ and $y$. The scores given by the voters to the candidates are the following, where $s_A(x)$ represents the score (on a 0-10 scale) voter A gives to candidate $x$: 
\begin{itemize}
\item $s_A(x)=6$, $s_A(y)=5$ 
\item $s_B(x)=8$, $s_B(y)=7$ 
\item $s_C(x)=1$, $s_C(y)=10$ 
\end{itemize}

The Condorcet-winner is candidate $x$ as both voters $A$ and $B$ prefer $x$ to $y$. But both $A$ and $B$ have a weak preference for $x$, and consider $y$ as almost acceptable as $x$, whereas $C$ has a strong preference for $y$ and strongly reject $x$. Even if candidate $y$ is the Condorcet-loser, it appears that electing $y$ instead of $x$ is a good compromise solution. Note that $y$ is the winner with any voting rule based on the use of the mean or the median score.

\end{ex}


The Deepest Voting is  a new family of social decision functions based on continuous evaluations. Let consider in the following that we have $n$ voters and $m$ candidates. Each voter can be seen as a point in $\mathbb{R}^m$ whose components are the graduations for each candidate. The set of all the voters' graduations is then a point cloud. The key idea of Deepest Voting is to consider the graduations of the {\it most central} voter of the cloud. This innermost (possibly imaginary) voter according to his graduations can be seen as the most representative of all the voters of the cloud, and his preferences should meet a large consensus among the others voters. The  social decision function simply gives the graduations of this innermost voter as output.

Quoting \citet{Serfling}: ``associated with a given distribution $F$ on $\mathbb{R}^m$ , a depth function is designed to provide a $F$-based center-outward ordering (and thus a ranking) of points $x$ in $\mathbb{R}^m$ . High depth corresponds to \emph{centrality}, low depth to \emph{outlyingness}". In other words, a depth function takes high (positive) values at the middle of a point cloud and vanishes out of it (see \citet{Zuo2000} for a rigorous definition of a depth function).
The very intuitive key idea of Deepest Voting is enriched by the large choice of the possible depth functions. There exists an infinity of ``center'' of a point cloud according to its definition.
Among others, the most famous examples are  halfspace depth by \citet{Tukey75}, simplicial depth by \citet{Liu1990OnAN} or weighted $L^p$ depths by \citet{Zuo2004}.

Note that $wL^pD(x)$, the weighted $L^p$ depth of a point $x$, given a cloud of $n$ points $\Phi(.,1),\hdots, \Phi(.,n)$ in $\mathbb{R}^m$ is defined by \citet{Zuo2004} as:

$$ w L^pD(x)=\frac{1}{1+\frac{1}{n} \sum_{j=1}^n \omega( \|\Phi(.,j)-x\|_p)}  $$ where $p>0$, $\omega$ is a non-decreasing and continuous function on $[0,\infty)$ with $\omega(\infty-)=\infty$  and $\| x-x' \|_p=\left(\sum_{j=1}^m |x_j-x'_j|^p\right)^{1/p}$.\\

If $\omega: x \rightarrow x^p$, then 

$$ w L^pD(x)=\frac{1}{1+\frac{1}{n} \sum_{j=1}^n   \sum_{i=1}^m |\Phi(i,j)-x_{i}|^p}.$$

Not considering tie-breaking procedure, the Majority Judgement (resp. the Range Voting) corresponds to the $wL^1$ Deepest Voting (resp. $wL^2$ Deepest Voting) with such $\omega$.

These methods are very promising and would take a great advantage of being explored by the of way of simulations. One can refer to a companion paper by \citet{aubin:hal-03192793} for details about Deepest Voting.

\section{The Impartial and Impartial Anonymous Culture models}\label{sec:ICIAC}

\subsection{IC/IAC models}

The Impartial Culture model (IC model) seems to be both the oldest and the most widely used simulation voting model. Introduced by \citet{Guilbaud1952}, IC model supposes that each preference order on the candidates is equally likely to be selected by each voter. As \citet{Diss2020} says,  ``this means that each individual randomly and
independently chooses his/her preference on the basis of a uniform probability
distribution across all linear (or weak) orders". The Impartial and Anonymous Culture (IAC) model was introduced  by \citet{GehrleinFishburn1976} and \citet{KugaNagatani1974}, and suppose that all preferences on the candidates concerning the whole set of voters are equally likely to appear. As \citet{Diss2020} noticed, ``both models are based on a notion of equi-probability, but the elementary events are preference profiles under IC and voting situations under IAC''. 

\subsection{Evaluations simulation IC models}

Both IC and IAC models are relevant for voting processes taking a set of preference profiles as input data. Evaluation-based voting processes (EBV process) need more than a set of preference profiles: EBV processes need a vector of numerical evaluations on the set of candidates.  Therefore an adaptation of both IC and IAC models is necessary to simulate evaluations.

IC model means that voters preferences are independent and identically distributed. An Evaluation Impartial Culture (E-IC) model should therefore mean that voters evaluations on each candidate are independent and identically distributed. Any distribution on $[0;1]$ should be used for the evaluations distribution (truncated normal or exponential distribution for instance) but without any other assumption on the distribution we suggest to focus on the uniform distribution on $[0;1]$.

\begin{defn}{\bf E-IC Uniform model}\\
 Evaluation Impartial Culture (E-IC) uniforme simulation model for the evaluation $e_{ij}$ of candidate $i$ by voter $j$ is defined as 
$$ \forall i \in 1,\ldots,d, \forall j \in 1,\ldots,n \quad   e_{ij} \sim \mathcal{U}[0;1] $$
\end{defn}

Setting the best candidate score to 1, and the worst candidate score  to 0 for each voter should be an option. The appropriate distribution is therefore a cumulative Dirichlet distribution, as stated in the following definition.

\begin{defn}{\bf E-IC Dirichlet model}\\
E-IC Dirichlet simulation model for the evaluation $e_{ij}$ of candidate $i$ by voter $j$ is defined as 
$$ \forall i \in 1,\ldots,d, \forall j \in 1,\ldots,n \quad   e_{i\sigma(j)}=\sum_{k=1}^{\sigma(j)} \delta_k $$
 with $(\delta_1,\ldots, \delta_d ) \sim Dir(1,\ldots,1)$ and $\sigma$  a random permutation on $\{1, \ldots,d\}$.
\end{defn}

IAC means that any voting situation should have the same probability to appear. In an evaluation perspective with a discrete scale of $k$ levels and $d$ candidates, there are $k^d$ different profiles (instead of $d!$ different preference orders). With a continuous scale there is an infinity of profiles. The fundamental difference between IC and IAC is that IAC doesn't suppose independence between voters. Therefore the use of multivariate copulas to simulate evaluation profiles should be interesting. In a nutshell  (see~\citet{nelsen2006} for a formal presentation of the subject) a copula is a multivariate cumulative distribution function which has all its margins 
uniformly distributed on the unit interval. 

Many copulas should be convenient to model evaluations of candidates by voters. We propose in the following to use Gaussian copulas to model dependences between evaluations.

\begin{defn}{\bf E-IAC Copula-based model}\\
E-IAC copula-based simulation model for the evaluation $e_{ij}$ of candidate $i$ by voter $j$ is defined as 
$$ \forall j \in 1,\ldots,n,\quad   (e_{1j}, \ldots, e_{dj}) \sim C(u_1, \ldots u_d)$$
where  $C$ is a multivariate specific copula and $u_1,\ldots, u_d \sim \mathcal{U}_{[0,1]}$.
\end{defn}

\subsection{Evaluations simulation: more realistic models}

The IC assumption is interesting in a theoretical point of view, but observing real-life voting situations shows that preferences over candidates are rarely identically distributed. In a more realistic point of view, we propose to weaken this condition: preference evaluations on each candidate are supposed to be distributed with the same basic probability distribution, but we suggest that the distribution parameters should randomly change from a candidate to another. We focus on two specific models, based on the use of a normal distribution or a Beta distribution.

\begin{defn}{\bf E-IC Normal model}\\
E-IC Normal simulation model for the evaluation $e_{ij}$ of candidate $i$ by voter $j$ is defined as 
$$ \forall i \in 1,\ldots,d, \forall j \in 1,\ldots,n \quad   e_{ij}=max(0, min(1, s_{ij})) \mbox{ with } s_{ij} \sim \mathcal{N}(\mu_i,\sigma_i)$$
where $\mathcal{N}(\mu_i,\sigma)$ is the Normal distribution with mean $\mu_i$ and standard deviation $\sigma_i$.
\end{defn}
Note that the obtained values should be transformed to be included in a $[0;1]$ interval. We suggest simply to cut values out of the interval to 0 or 1.  

\begin{defn}{\bf E-IC Beta model}\\
E-IC Beta simulation model for the evaluation $e_{ij}$ of candidate $i$ by voter $j$ is defined as 
$$ \forall i \in 1,\ldots,d, \forall j \in 1,\ldots,n \quad   e_{ij}=s_{ij} \mbox{ with } s_{ij} \sim B(\alpha_i,\beta_i)$$
where $B(\alpha_i,\beta_i)$ is the Beta distribution of parameters $\alpha_i$ and $\beta_i$.
\end{defn}

For E-IAC model the dependence parameters (\emph{i.e.} the covariance matrix) should be given. The more simple is to choose a unique correlation coefficient for all pairs of evaluation variables, but we can also use random correlation coefficient for all pairs\footnote{Note that the random correlation matrix should be positive semi-definite.}. Another approach, in the same spirit as \citet{CugliariRolland2018}, consists in learning the copula (or at least the coefficients of a given standard copula, as the Gaussian one) from real existing data.

\subsection{Links between IC and E-IC models}

Evaluation simulations based on models E-IC are coherent with IC models for preferences simulations as far as evaluations $e(i,j)$ are IID. Models E-IC Uniform and E-IC Dirichlet (due to the permutation) are IID by construction. Models E-IC Normal and Beta are IID as far as the parameters ($\mu$, $\sigma$, $\alpha$ or $\beta$) are the same for each candidate. 
As far as evaluations $e(i,j)$ are IID, these simulations can be used also to obtain preference orders simulations that follow the IC model.

\section{Spatial voting simulations}\label{sec:simuspatiale}

\subsection{Downsian voting model}

Following the early work of \citet{Downs1957}, spatial voting simulations have been developped as intuitionistic voting models. This model is based on the use of an euclidean distance between the candidates and the voters described in the same uni- or multi-dimensional space: the smaller the distance, the greater the preference.   \citet{Tideman2010} show that a spatial model ``describes the observations in data sets much more accurately'' than other models. This approach is very efficient to simulate real-like voting data (especially when the political scene is strongly polarized) and therefore to calculate precise probabilities of voting situations or paradox to happen.

\subsection{Spatial evaluation simulation}

As spatial model is directly built on a distance as evaluation, it is very easy to turn it into an evaluation simulation. Voters $v_1, \ldots, v_n$ and candidates $c_1, \ldots, c_d$ are randomly uniformly generated as points inside the $[0;1]^k$ hypercube, where  $k$ is a given dimension parameter. Parameter $k$ should be seen as the number of dimensions needed to analyse the political opinions of the candidates  (typically $k=2$ or $3$, see \citet{Armstrong2020} to investigate the choice of $k$).

\begin{defn}{\bf Spatial simulation model}\\
Spatial simulation model for the evaluation $e_{ij}$ of candidate $i$ by voter $j$ is defined as 
$$ \forall i \in 1,\ldots,d, \forall j \in 1,\ldots,n \quad   e_{ij}=f(d(c_i,v_j))$$
where $d$ is a distance between $c_i$ and $v_j$ and $f$ a non-increasing function mapping $\mathbb{R^+}$ to $[0,1]$.
\end{defn}
Typically, the simplest spatial simulation model is
$ \forall i \in 1,\ldots,d, \forall j \in 1,\ldots,n$ $e_{ij}=max(0,(1-d(c_i,v_j)))$ with $d$ the euclidean distance. Other functions $f$ are possible, as for example the sigmoïd: $e_{ij}=(1+e^{\lambda(\beta d(c_i, v_j)-1)})^{-1}$, $\lambda>0$ and $\beta>0$.

\section{Models comparison}\label{sec:compmethods}

In this section, we investigate the differences between the proposed  models by the use of several experiments.

\subsection{Experiments}

Sections \ref{sec:ICIAC} and \ref{sec:simuspatiale} introduced theoretical simulation models, which often need one or several parameters to be used in practice.
\begin{itemize}
\item Model E-IC-Uniform and model E-IC-Dirichlet don't need any parameter.
\item Model E-IAC (Copula-based) needs the choice of a specific copula. We used the multivariate Gaussian copula to experiment the model, for its good trade-off between simplicity of use and complexity of interactions modelling.
\item  Model E-IC-Normal needs two parameters, mean and standard deviation. For each candidate the mean is randomly chosen uniformly in $[0,1]$, whereas the standard deviation is always the same and fixed at 0.25, in order to ensure the same shape of evaluations for all candidates.
\item Model E-IC-Beta needs two parameters $\alpha$ and $\beta$. Experiments lead us to randomly choose for each candidate a pair $(\alpha, \beta)$ by the following process: for each parameter, choose with the probability $1/2$ either a value obtained through a uniform distribution on $[0.5,1]$ or  a value obtained through a uniform distribution on $[1,5]$. Each configuration $(\alpha< 1; \beta < 1)$, $(\alpha< 1; \beta > 1)$, $(\alpha> 1; \beta < 1)$, $(\alpha> 1; \beta > 1)$ is equally likely.
\item Spatial model has been experimented on 2 dimensions (for less than 5 candidates) and 3 dimensions (for mode than 5 candidates). Parameter $\lambda$ controls the shape of the decreasing curve, whereas parameter $\beta$ controls the position of the decreasing curve. Experiments have been made with  $\lambda=5$ and $\beta=2$.
\end{itemize}

\begin{center}
\begin{figure}
\begin{center}
\includegraphics[width=\textwidth]{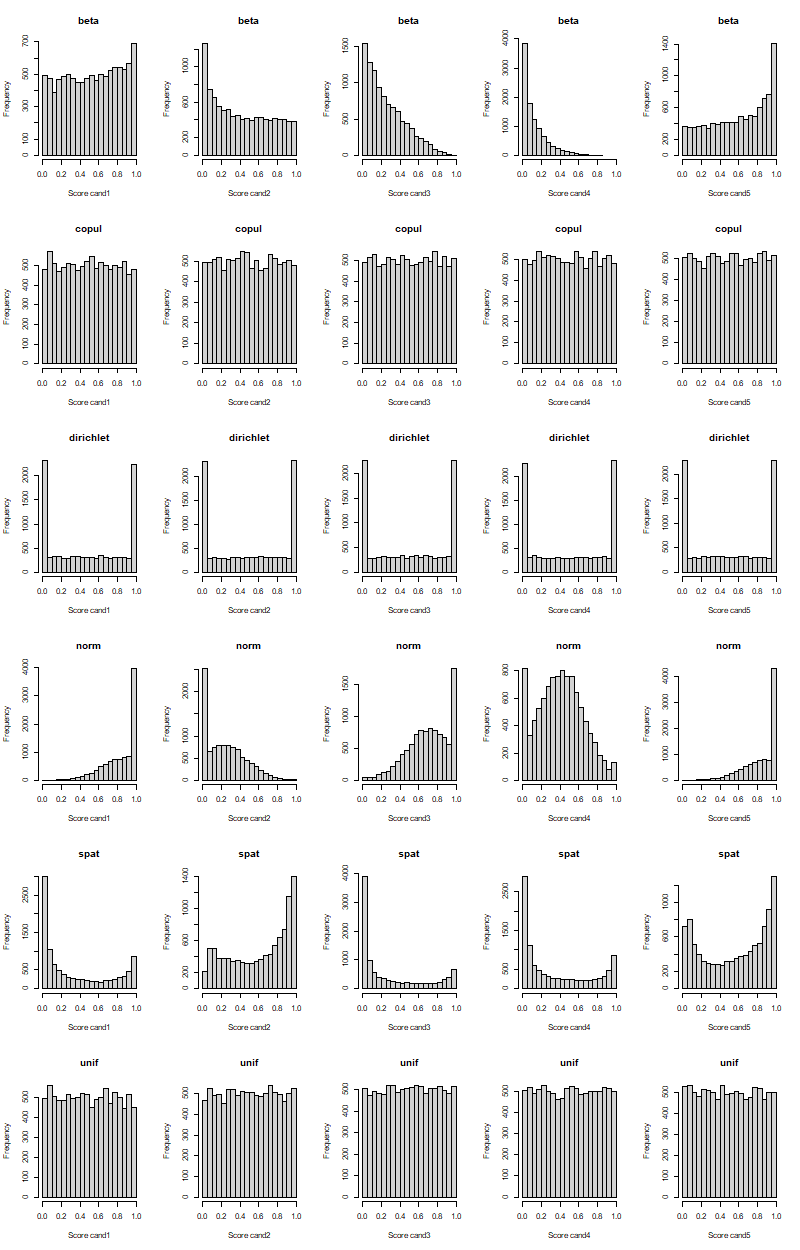}
\caption{Evaluations histograms for 5  candidates and 6 methods \label{fig:6methodes}}
\end{center}
\end{figure}
\end{center}

\begin{center}
\begin{figure}
\begin{center}
\includegraphics[width=\textwidth]{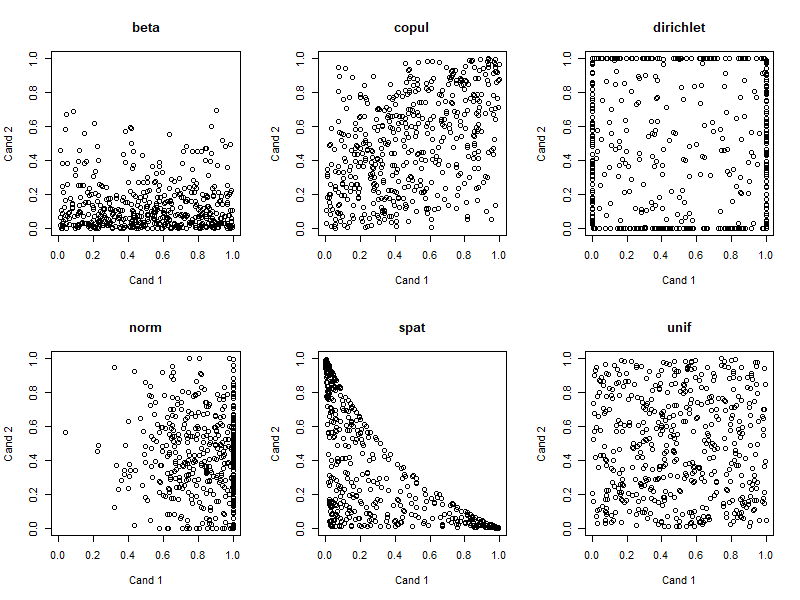}
\caption{Example of bi-dimensionnal plot for cross evaluations of 2 candidates for each simulation model \label{fig:plot2D}}
\end{center}
\end{figure}
\end{center}

Figure \ref{fig:6methodes} presents the histograms of simulations with 5 candidates for each of the 6 methods.
Figure \ref{fig:plot2D} shows an example of relation between the scores obtained by two different candidates (each point represents a voter). This figure emphasis the difference between each simulation method and the uniform distribution. Copula-based method permits to control the correlation between the two evaluations. Dirichlet is an uniform distribution except it forces at least one candidate to score at 1 and another candidate to score at 0. Therefore for 5 candidates each of them scores to 0 approximately for 20\% of the voters and to 1 another 20\%. E-IC Normal model over-represents 0 and 1 scores due to truncation. E-IC  Beta model under-represents 0 and 1 scores. The spatial model leads to very specific distribution, due to the structure of the unit cube: there is no point which is both far from candidate~1 and candidate~2.

\subsection{Comparison with real-world data}
Only few real-world data are available in the framework of EBV rules. We propose to study data about the 2009 European Election Survey carried out in Denmark and presented in  \citet{Denmark} and \citet{Armstrong2020}, and a french experiment about the 2017 presidential election conducted by \citet{bouveret_sylvain_2018_1199545}. Are our models able to simulate real data? It is important not to be overconfident, as we really don't know if the observed data distribution are common or very specific. It should be very interesting to make more experiments in order to have strong assumption of a ``classical'' evaluation distribution.

\begin{center}
\begin{figure}
\begin{center}
\includegraphics[width=\textwidth]{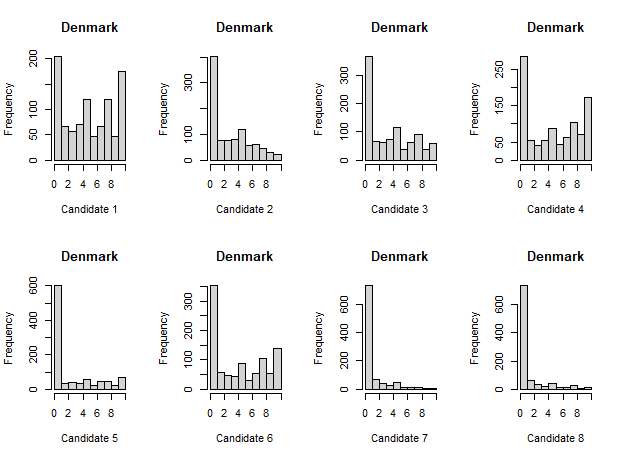}
\caption{Evaluation histograms for 8 candidates - Denmark Survey \label{fig:Denmark}}
\end{center}
\end{figure}
\end{center}

\begin{center}
\begin{figure}
\begin{center}
\includegraphics[width=\textwidth]{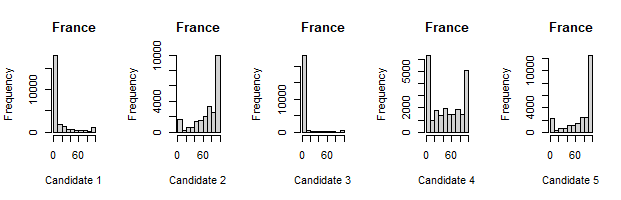}
\caption{Evaluation histograms for 5 candidates - French presidential election \label{fig:France}}
\end{center}
\end{figure}
\end{center}

\begin{center}
\begin{figure}
\begin{center}
\includegraphics[width=\textwidth]{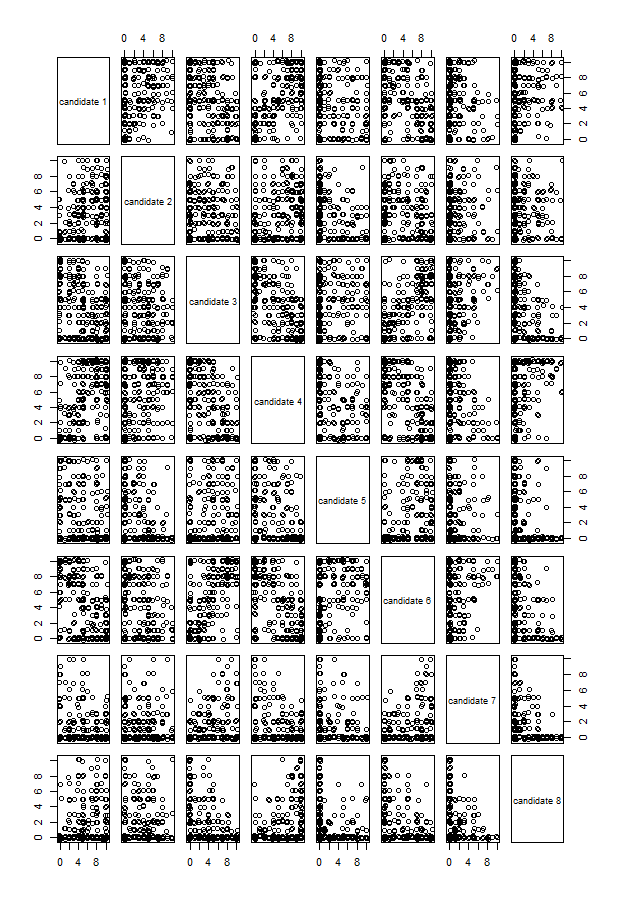}
\caption{Bi-dimensionnal plot for cross evaluations of all the pairs of candidates - Denmark survey  \label{fig:Denmark2D} - note that the data have been slightly corrupted to emphasize the density}
\end{center}
\end{figure}
\end{center}

\begin{center}
\begin{figure}
\begin{center}
\includegraphics[width=\textwidth]{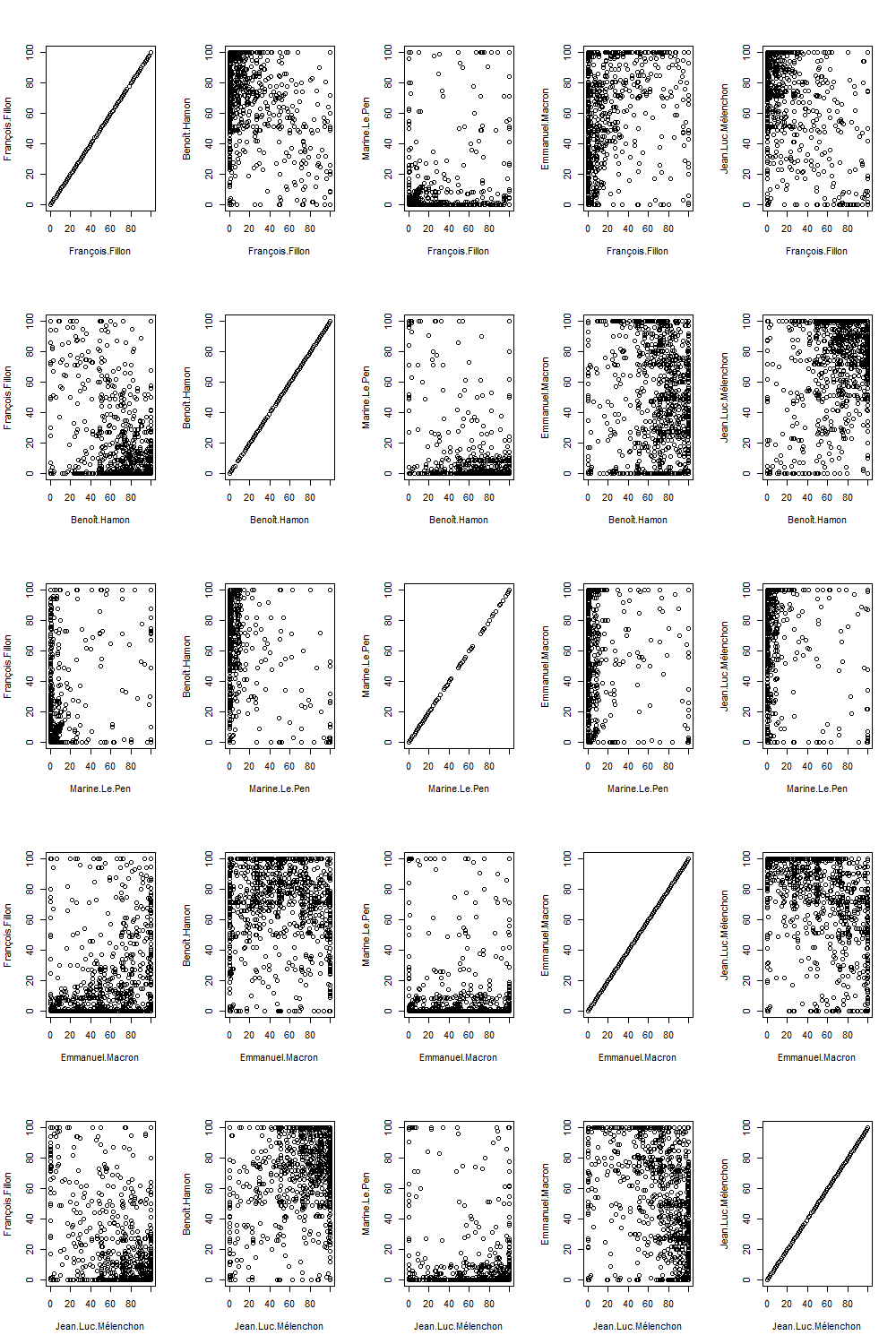}
\caption{Bi-dimensionnal plot for cross evaluations of all the pairs of candidates - French presidential election \label{fig:pres_candidats_croises}}
\end{center}
\end{figure}
\end{center}

The first database (Denmark European Election Survey) includes evaluations on a discrete 0-10 scale of 8 candidates by 972 voters.
Evaluations histograms for each candidate is shown on figure \ref{fig:Denmark} and bi-dimensional plots for all pairs of candidates on figure \ref{fig:Denmark2D}. The second database (French presidential election 2017) includes 26633 voters and has been limited to the 5 best candidates.  Figure \ref{fig:France} shows the evaluations histograms for each candidate, and figure \ref{fig:pres_candidats_croises} bi-dimensional plots for all pairs of candidate.

The aim of simulations we processed was not to fit the data of both Denmark and France database. However it is interesting to have a look on the shape of evaluations histograms and bi-dimensional plots to see if simulation models lead to plausible data or not.
Both E-IC uniform model and E-IAC copula model seem far away from real evaluations, as they minimize the strong polarisation of the voters' opinion. The uniform assumption seems to be unrealistic, even if it is supposed to be the only one we can suppose without any information on the real distribution, as in IC models.
E-IC Normal model seems more realistic, as it produces peaks in 0 and/or 1 due to the truncations of the distribution. However, it overestimates moderate evaluations. E-IC Dirichlet model seems better as it takes into account the importance of 0 and 1 evaluations and does not overestimate intermediate scores. However this model leads to give the same proportion of 0 and 1 to each candidate, which seems unrealistic. Both E-IC Beta and spatial models seem to better capture the general shape of real evaluations, as intermediates scores are not uniform. Spatial model can also capture the fact that some candidates are strongly polarized, with both numerous 0 and 1 evaluations.

\subsection{Conclusive recommendation}

We introduced in this paper several models to simulate evaluation-based voting data in a probabilistic-based analysis perspective of EBV rules. Some of them are parametric (E-IAC Copula-based, E-IC Normal, E-IC Beta, spatial models) and therefore seem useful to fit observed data. The E-IC uniform model should appear to be very legitimate without any assumption on the voters and candidates, but the obtained scores seem unrealistic. Therefore, the frequency of appearance estimations for voting paradox obtained with such simulations should also be unrealistic. Spatial model seems also legitimate to represent the proximity between candidates and voters. The obtained simulation are more realistic than the one obtained with the uniform model. Note that even if the E-IC Beta model leads to individual evaluation distributions similar to these obtained with spatial model, it differs drastically from the spatial model in terms of  relations between scores of two candidates. We therefore recommend to use a spatial model for evaluations simulations in a evaluation-based voting scheme.

\bibliographystyle{spbasic} 
\bibliography{Bibliosimu}

\begin{thebibliography}{28}
\providecommand{\natexlab}[1]{#1}
\providecommand{\url}[1]{{#1}}
\providecommand{\urlprefix}{URL }
\expandafter\ifx\csname urlstyle\endcsname\relax
  \providecommand{\doi}[1]{DOI~\discretionary{}{}{}#1}\else
  \providecommand{\doi}{DOI~\discretionary{}{}{}\begingroup
  \urlstyle{rm}\Url}\fi
\providecommand{\eprint}[2][]{\url{#2}}

\bibitem[{Armstrong et~al.(2020)Armstrong, Bakker, Carroll, Poole, and
  Rosenthal}]{Armstrong2020}
Armstrong D, Bakker R, Carroll C Rand~Hare, Poole K, Rosenthal H (2020)
  Analyzing Spatial Models of Choice and Judgment (2nd ed.). CRC Press

\bibitem[{Aubin et~al.(2021)Aubin, Gannaz, Leoni-Aubin, and
  Rolland}]{aubin:hal-03192793}
Aubin JB, Gannaz I, Leoni-Aubin S, Rolland A (2021) {Deepest Voting: a new way
  of electing}, \urlprefix\url{https://hal.archives-ouvertes.fr/hal-03192793},
  working paper or preprint

\bibitem[{Balinski and Laraki(2007)}]{Balinski2007}
Balinski M, Laraki R (2007) A theory of measuring, electing and ranking.
  Proceedings of the National Academy of Sciences USA 104(21):8720--8725

\bibitem[{Balinski and Laraki(2011)}]{Balinski2011}
Balinski M, Laraki R (2011) Majority Judgment; Measuring, Ranking, and
  Electing. MIT Press

\bibitem[{Balinski and Laraki(2020)}]{Balinski2020}
Balinski M, Laraki R (2020) Majority judgment vs. majority rule. Social Choice
  and Welfare 54(2):429--461

\bibitem[{Bouveret et~al.(2018)Bouveret, Blanch, Baujard, Durand, Igersheim,
  Lang, Laruelle, Laslier, Lebon, and Merlin}]{bouveret_sylvain_2018_1199545}
Bouveret S, Blanch R, Baujard A, Durand F, Igersheim H, Lang J, Laruelle A,
  Laslier JF, Lebon I, Merlin V (2018) Voter autrement 2017 - online
  experiment. \doi{10.5281/zenodo.1199545},
  \urlprefix\url{https://doi.org/10.5281/zenodo.1199545}

\bibitem[{Brams and Fishburn(2007)}]{Brams2007}
Brams S, Fishburn PC (2007) Approval voting. Springer

\bibitem[{Cugliari and Rolland(2018)}]{CugliariRolland2018}
Cugliari J, Rolland A (2018) Simulation of multicriteria data. EURO Journal on
  Decision Processes 6:21--37

\bibitem[{Diss and Kamwa(2020)}]{Diss2020}
Diss M, Kamwa E (2020) Simulations in models of preference aggregation.
  Œconomia 10(2):279--308

\bibitem[{Downs(1957)}]{Downs1957}
Downs A (1957) An economic theory of political action in a democracy. Journal
  of Political Economy 65(2):135--150

\bibitem[{Fabre(2021)}]{Fabre2021}
Fabre A (2021) Tie-breaking the highest median: alternatives to the majority
  judgment. Social Choice and Welfare 56(1):101--124

\bibitem[{Felsenthal and Machover(2012)}]{Felsenthal2012}
Felsenthal DS, Machover M (2012) Electoral Systems; Paradoxes, Assumptions, and
  Procedures. Springer

\bibitem[{Gehrlein and Fishburn(1976)}]{GehrleinFishburn1976}
Gehrlein WV, Fishburn PC (1976) Condorcet’s paradox and anonymous preference
  profiles. Public Choice 26(1):1--18

\bibitem[{Green-Armytage et~al.(2016)Green-Armytage, Tideman, and
  Cosman}]{Green-Armytage2016}
Green-Armytage J, Tideman T, Cosman R (2016) Statistical evaluation of voting
  rules. Social Choice and Welfare 46(1):183--212

\bibitem[{Guilbaud(1952)}]{Guilbaud1952}
Guilbaud GT (1952) Les théories de l'intérêt général et le problème
  logique de l'agrégation. Economie Appliquée 5(4):501--584

\bibitem[{Kuga and Hiroaki(1974)}]{KugaNagatani1974}
Kuga K, Hiroaki N (1974) Voter antagonism and the paradox of voting.
  Econometrica 42(6):1045--1067

\bibitem[{Liu(1990)}]{Liu1990OnAN}
Liu R (1990) On a notion of data depth based on random simplices. Annals of
  Statistics 18:405--414

\bibitem[{Liu et~al.(2006)Liu, Serfling, and Souvaine}]{Serfling}
Liu RY, Serfling R, Souvaine DL (eds) (2006) Data Depth: Robust Multivariate
  Analysis, Computational Geometry and Applications, Proceedings of a {DIMACS}
  Workshop, New Brunswick, New Jersey, USA, May 14-16, 2003, {DIMACS} Series in
  Discrete Mathematics and Theoretical Computer Science, vol~72, {DIMACS/AMS},
  \doi{10.1090/dimacs/072}, \urlprefix\url{https://doi.org/10.1090/dimacs/072}

\bibitem[{Nelsen(2006)}]{nelsen2006}
Nelsen R (2006) An Introduction to Copulas, 2nd edn. Springer

\bibitem[{Plassmann and Tideman(2014)}]{Plassman2014}
Plassmann F, Tideman T (2014) {How frequently do different voting rules
  encounter voting paradoxes in three-candidate elections?} Social Choice and
  Welfare 42(1):31--75

\bibitem[{Smith(2000)}]{Smith00rangevoting}
Smith WD (2000) Range voting

\bibitem[{Tideman and Plassmann(2010)}]{Tideman2010}
Tideman TN, Plassmann F (2010) The structure of the election-generating
  universe, unpublished

\bibitem[{Tideman and Plassmann(2012)}]{Tideman2012}
Tideman TN, Plassmann F (2012) Developing the empirical side of computational
  social choice. In: International Symposium on Artificial Intelligence and
  Mathematics, {ISAIM} 2012, Fort Lauderdale, Florida, USA, January 9-11, 2012

\bibitem[{Tideman and Plassmann(2014)}]{Tideman2014}
Tideman TN, Plassmann F (2014) Which voting rule is most likely to choose the
  "best" candidate? Public Choice 158(3/4):331--357

\bibitem[{Tukey(1975)}]{Tukey75}
Tukey JW (1975) Mathematics and the picturing of data. In: Proceeding of the
  International Congress of Mathematicians, Vancouver 1974, Canadian
  Mathematical Congress, Montreal, vol~2, pp 523--531

\bibitem[{{van Egmond} et~al.(2013){van Egmond}, {van der Brug}, Hobolt,
  Franklin, and Sapir}]{Denmark}
{van Egmond} M, {van der Brug} W, Hobolt S, Franklin M, Sapir EV (2013)
  European Parliament Election Study 2009, Voter Study. GESIS Data Archive,
  Koln

\bibitem[{Zuo(2004)}]{Zuo2004}
Zuo Y (2004) Robustness of weighted {L}p–depth and {L}p–median. Allgemeines
  Statistisches Archiv 88(2):215--234

\bibitem[{Zuo and Serfling(2000)}]{Zuo2000}
Zuo Y, Serfling R (2000) General notions of statistical depth function. Annals
  of Stat 28:461--482

\end{thebibliography}

\end{document}